\documentclass[letterpaper]{article} 
\usepackage{aaai2026}  
\usepackage{times}  
\usepackage{helvet}  
\usepackage{courier}  
\usepackage[hyphens]{url}  
\usepackage{graphicx} 
\usepackage{natbib}  
\usepackage{caption} 
\usepackage{algorithm}
\usepackage{algorithmic}

\usepackage{newfloat}
\usepackage{listings}
\DeclareCaptionStyle{ruled}{labelfont=normalfont,labelsep=colon,strut=off} 
\floatstyle{ruled}
\newfloat{listing}{tb}{lst}{}
\floatname{listing}{Listing}
\usepackage{latexsym}
\usepackage{amssymb}
\usepackage{amsmath}
\usepackage{amsthm}
\usepackage{booktabs}
\usepackage{enumitem}
\usepackage{graphicx}
\usepackage{array}
\newcolumntype{P}[1]{>{\raggedright\arraybackslash}p{#1}}
\newcolumntype{C}[1]{>{\centering\arraybackslash}p{#1}}

\title{Reasoning Transfer for an Extremely Low-Resource and Endangered Language: Bridging Languages Through Sample-Efficient Language Understanding}
\author{
    Khanh-Tung Tran\thanks{Corresponding author.}, Barry O’Sullivan, Hoang D. Nguyen
}
\affiliations{
    Research Ireland Centre for Research Training in Artificial Intelligence and Insight Research Ireland Centre for Data Analytics\\ School of Computer Science and Information Technology, University College Cork, Ireland\\
    123128577@umail.ucc.ie, b.osullivan@cs.ucc.ie, hn@cs.ucc.ie
}

\usepackage{bibentry}

\begin{document}

\maketitle

\begin{abstract}
Recent advances have enabled Large Language Models (LLMs) to tackle reasoning tasks by generating chain-of-thought (CoT) rationales, yet these gains have largely applied to high-resource languages, leaving low-resource languages 
behind.
In this work, we first investigate CoT techniques in extremely low-resource scenarios through previous prompting, model-editing, and fine-tuning approaches.
We introduce \emph{English-Pivoted CoT Training}, leveraging the insight that LLMs internally operate in a latent space aligned toward the dominant language. Given input in a low-resource language, we perform supervised fine-tuning to generate CoT in English and output the final response in the target language. Across mathematical reasoning benchmarks, our approach outperforms other baselines with up to 28.33\% improvement in low-resource scenarios. Our analysis and additional experiments, including Mixed-Language CoT and Two-Stage Training, show that explicitly separating language understanding from reasoning enhances cross-lingual reasoning abilities. To facilitate future work, we also release \emph{LC2024}, the first benchmark for mathematical tasks in Irish, an extremely low-resource and endangered language. Our results and resources highlight a practical pathway to multilingual reasoning without extensive retraining in every extremely low-resource language, despite data scarcity.
\end{abstract}

\begin{links}
    \link{Code, Dataset, Appendix}{https://github.com/ReML-AI/english-pivoted-cot}
\end{links}

\section{Introduction}
    Large Language Models (LLMs) have showcased exceptional performance in a wide-range of  task~\cite{zhao2025surveylargelanguagemodels,Zhang2024}, particularly in English, due to the vast amount of available data during pre-training~\cite{qin2024multilingual,DBLP:conf/ecai/Guo0RLZZ23,Kim2024}. Recent advances in post-training techniques, such as reasoning scaling~\cite{openai2024openaio1card}, which explicitly trains models to generate chain-of-thought (CoT) reasoning, have significantly enhanced model accuracy in tasks such as mathematics and programming. However, such advances have unintentionally widened the performance gap between popular languages like English and low, extremely low-resource languages~\cite{ji2025test}. This disparity arises primarily because post-training methods typically require high-quality, manually curated datasets, predominantly available only in English, and misalignment and inherent biases within multilingual training corpora~\cite{ghosh2025multilingualmindsurvey}.

    \begin{figure*}[ht]
        \centerline{\includegraphics[width=1.0\linewidth]{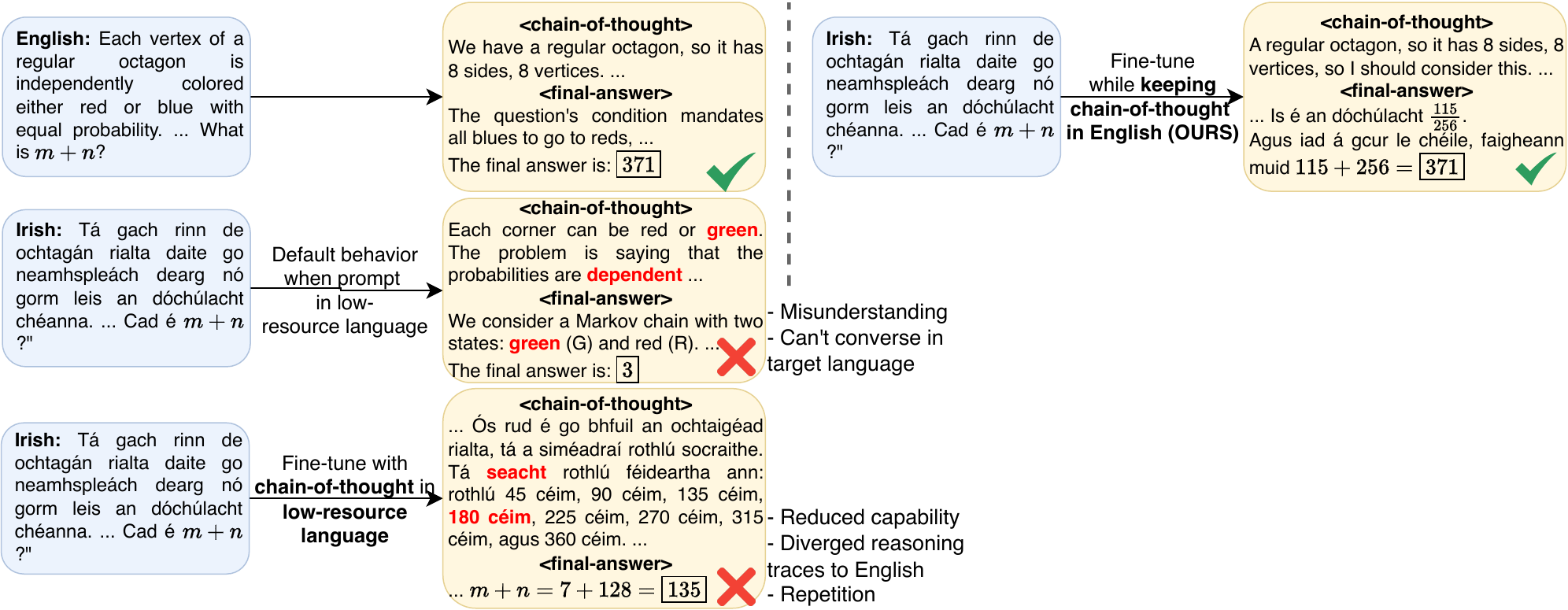}}
        \caption{Illustrative example of model behavior (r1-distill-Llama-8B) when prompted with the same problem in English (robust reasoning) versus a low-resource language - Irish (reduced understanding and conversational ability). Training with an Irish chain-of-thought diverges from baseline, while training with English chain-of-thoughts achieves the best of both worlds.}
        \label{fig:example}
    \end{figure*}

    Figure \ref{fig:example} illustrates this phenomenon by comparing outputs from a small-size state-of-the-art reasoning model (r1-distill-Llama-8B~\cite{deepseekai2025deepseekr1incentivizingreasoningcapability}) given the same mathematical problem presented in English and in Irish, classified endangered by UNESCO~\cite{Unesco2010-st}, less than 0.01\% appearance in Common Crawl~\cite{commoncrawl}. When the question is presented in English, the model reasoning process can lead to correct solution, however, with the same problem in Irish, the model misunderstands, unable to converse in target language, causing the CoT process to fail and resulting in an incorrect answer. Our proposed solution addresses this issue by aligning the model’s reasoning process across languages, allowing it to ``think'' internally in its dominant language (English).

    Enabling CoT reasoning in limited-resource languages is critical for developing AI systems that function effectively across diverse linguistic contexts.
    Despite its importance, this remains a relatively unexplored domain, leaving low-resource and extremely low-resource languages underrepresented. Current low-resource CoT techniques can be split into three broad categories: (1) multilingual fine-tuning~\cite{lai-nissim-2024-mcot,anand2024multilingualmathematicalreasoningadvancing}, (2) prompting strategies (translation-based~\cite{shi2023language} or self alignment-based prompting~\cite{qin-etal-2023-cross}), and (3) model editing (model merging~\cite{yadav2023tiesmerging,tao-etal-2024-unlocking}, layer swapping~\cite{bandarkar2025layer}).
    In general, while these methods show promise, they often require extra training data, additional modules, or incur translation errors, limiting their practical usability in truly low-resource settings.

    Inspired by recent advances in reasoning scaling, where models explicitly generate CoT traces without the need for carefully crafted reasoning prompts, and recent findings on the dynamics of language models across layers~\cite{shi2023language,wendler-etal-2024-llamas,tran-etal-2024-irish} (where distinct layers/neurons specialize in language understanding and others in reasoning, typically biased towards English), we explore the connections between reasoning and multilinguality, particularly how reasoning can be transferred across languages by keeping the explicit CoT in English, while boosting the low-resource language understanding and generation of the model while keeping inputs and final responses in the target language. Our method allows the LLM to interpret a problem in the target language, perform CoT in English, where it is most robust, and then subsequently generate accurate final responses in the target language. Our contributions are summarized as follows:
    \begin{itemize}
        \item \emph{English-Pivoted CoT Training}, a novel and effective method for adapting LLM’s reasoning capabilities to extremely low-resource scenarios. Our approach explicitly aligns reasoning process across languages, addressing linguistic misalignment and boosting performances.
        \item Our results demonstrate successful transfer of reasoning capability to Irish, an extremely low-resource language, by leveraging model’s internal reasoning representations across languages, upto 28.33\% improvement.
        \item Our analysis and ablations, including generalization to medium and high-resource languages, mixed CoT language and two-stage training further highlight the benefit of separating language understanding and reasoning, demonstrating a practical pathway for multilingual reasoning without extensive retraining in each language.
        \item \emph{LC2024}, the first-ever dataset for evaluating mathematical reasoning in Irish, an endangered language.
    \end{itemize}

\section{Related Works}
    \subsection{CoT Reasoning for Low-Resource Languages}
        Reasoning is formally described as the cognitive process of logically analyzing available information to reach conclusions, enabling both humans and AI systems to address complex problems~\cite{ghosh2025multilingualmindsurvey}. 
        Chain-of-thought~\cite{wei2022chain,zelikman2022star} has emerged as a powerful technique for improving reasoning in LLMs by generating explicit step-by-step rationales.
        Recent works explore CoT in low-resource languages~\cite{shi2023language,qin-etal-2023-cross}.
        \cite{shi2023language} finds that few-shot prompting with CoT in English consistently achieves competitive performance to prompting in the native language. To bridge the gap between languages, alignment techniques have been carried out to align representation between low-resource languages and English through learning with parallel sentences~\cite{zhu-etal-2024-question,li-etal-2024-improving-context} or additional multilingual encoders~\cite{huang2024mindmerger}. 
        On the other hand, there are also attempts to directly fine-tune LLMs in reasoning tasks across multiple languages, by obtaining reasoning data in low-resource languages, mostly through neural machine translation~\cite{anand2024multilingualmathematicalreasoningadvancing,lai-nissim-2024-mcot}.

        While prior works highlight stronger performance when aligning language models to reason in English for tasks in low-resource languages, they typically leverage prompting techniques or require additional modules or parallel corpus. In this work, we explicitly train LLMs to separate their reasoning and language understanding capabilities by separating the languages of internal CoT and final response.

        \begin{figure*}[ht]
        \centering
        \includegraphics[width=0.85\linewidth]{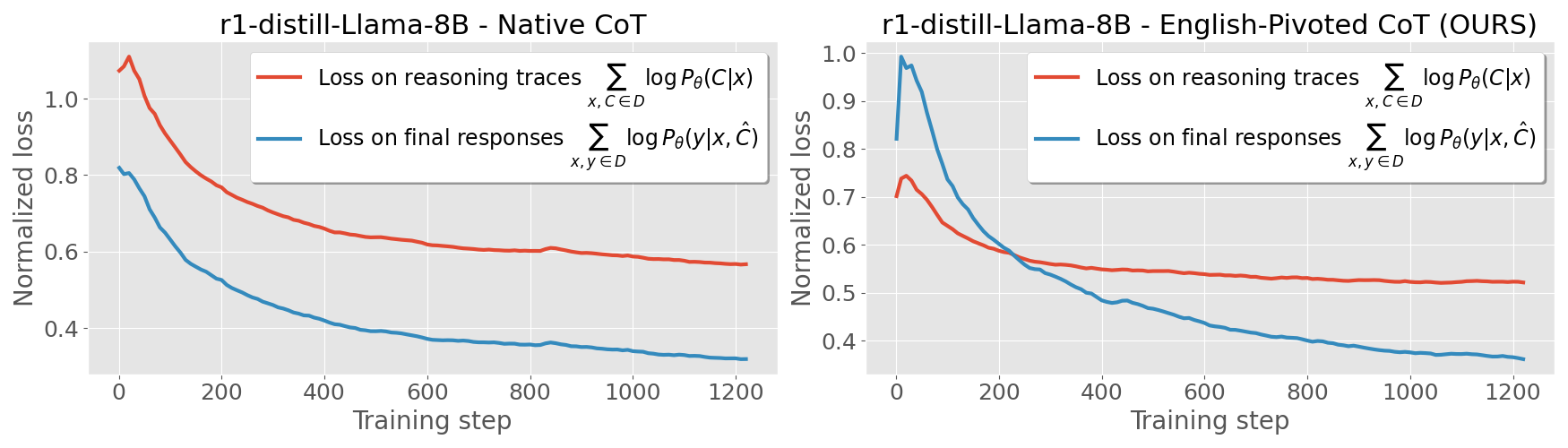}
        \caption{Loss curves over the training process for Native CoT Training (Left) and our approach, English-Pivoted CoT Training (right), normalized with exponential moving average with a smoothing weight of 0.95. Our method shows lower initial reasoning loss and slower decline, indicating effective separation of English reasoning from target-language responses.}
        \label{fig:loss_comparison}
        \end{figure*}

    \subsection{Alignment of Internal Representations in Multilingual LLMs}
        There has been a large interest in investigating and aligning how multilingual LLMs organize and share knowledge between languages internally. A key question is whether LLMs ``think'' in English, or rely on English-centric representations or rationales even when operating in other languages. There is evidence that multilingual models often map other languages into an implicit English latent space~\cite{li2024align,zhu-etal-2024-question}.
        Recent works analyze intermediate embeddings across transformer layers and found distinct phases of operation, where middle layers perform reasoning in a latent space steer toward dominant language (English)~\cite{wendler-etal-2024-llamas,tran-etal-2024-irish}.


        These alignment efforts~\cite{tao-etal-2024-unlocking,bandarkar2025layer} seek to improve cross-lingual generalization, allowing knowledge learned in one language to transfer to others. Our work is informed by these studies but takes a distinct direction. While prior works often relies on external models or separate bilingual modules, our approach aligns the reasoning process within the model’s own representations. This allows us to transfer reasoning skills to low-resource languages without retraining from scratch, advancing the understanding of how multilingual LLMs can maintain consistent reasoning across diverse languages.

\section{English-Pivoted CoT Training for Reasoning Transfer}

    We introduce \emph{English-Pivoted CoT Training}, leveraging the model’s robust reasoning capabilities developed predominantly in English, by explicitly constraining the intermediate CoT reasoning steps to English, while maintaining inputs and final responses in the target low-resource language. Formally, let $x$ denote an input problem expressed in the target (low-resource) language, and let $y$ represent the final answer also in the target language. We denote the chain-of-thought reasoning trace as $C$, which is constrained to be in English. The rationale behind this design is that models typically have been extensively trained in English for reasoning tasks, and we hypothesize that their latent spaces are more robustly aligned with English reasoning processes. Our method is designed to learn the conditional probability distribution:

        \begin{figure*}[ht]
        \centering
        \includegraphics[width=1.0\linewidth]{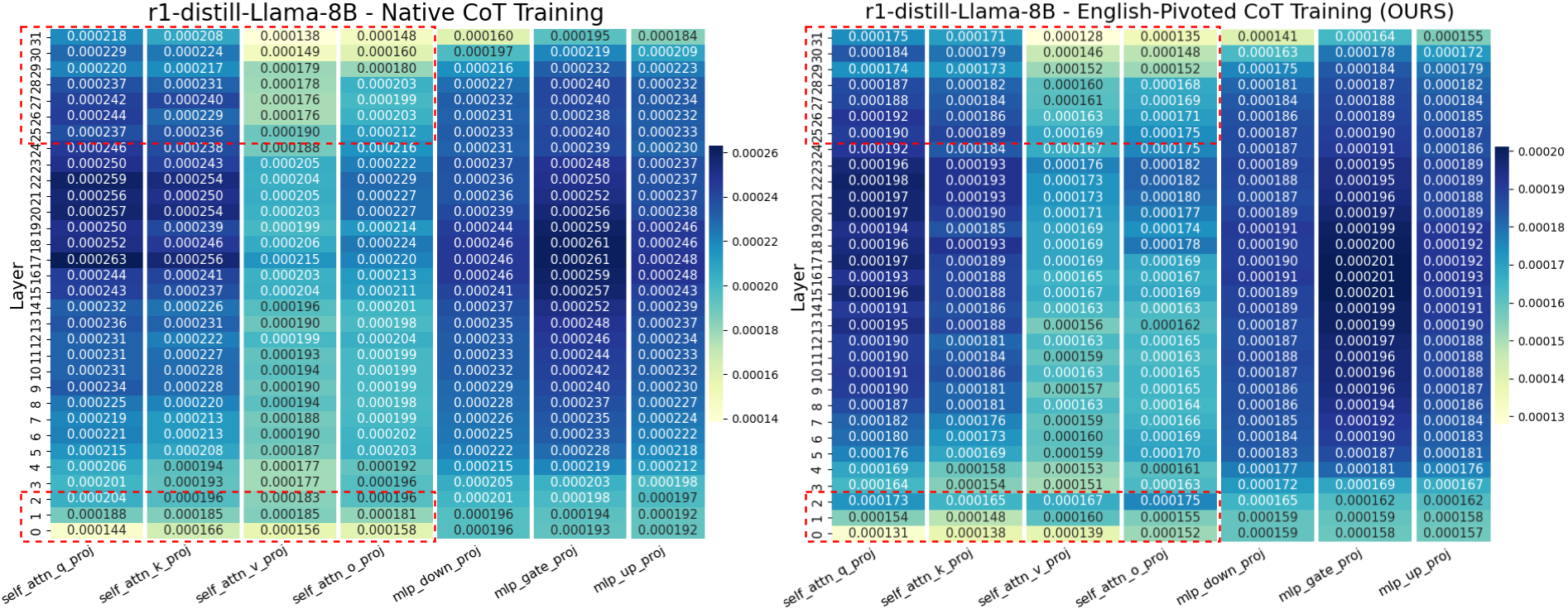}
        \caption{Parameter updates (mean absolute differences) of Native CoT Training (Left) and English-Pivoted CoT Training (right) for r1-distill-Llama-8B. English-Pivoted CoT Training focuses more on language comprehension layers (red boxes).}
        \label{fig:weight_changes}
        \end{figure*}

    \begin{equation}
        P_{\theta}(C, y \vert x) = P_\theta(C \vert x) P_\theta(y|x,C)
    \end{equation}
    where $\theta$ represents the parameters of the language model, $C=(s_1, \dots, s_n)$ is the chain-of-thought, comprising natural language reasoning steps $s_i$. Following recent works~\cite{zelikman2022star,openaio1}, we explicitly separate $C$ and $y$ by a special token (e.g. \emph{``$<$/think$>$''}) to mark the transition between the reasoning trace and the final answer.

    Given a training dataset $D=\{(x, C, y)\}$ consisting of tuples where:
    \begin{itemize}
        \item $x \in S_T$: input in the target low-resource language.
        \item $C \in S_{en}$: reasoning trace in English.
        \item $y \in S_T$: final answer in the target low-resource language.
    \end{itemize}
    we optimize the following objective during training:
    {\small	\begin{align}
        \mathcal{L}(\theta) & = \sum_{x, C, y \in D} \log P_\theta(C, y \vert x) \\
        & = \sum_{x, C, y \in D}[\log P_\theta(C \vert x) + \log P_\theta(y \vert x, \hat{C})] \\
        & = \sum_{x, C \in D}\log P_\theta(C \vert x) + \sum_{x, y \in D} \log P_\theta(y \vert x, \hat{C}) \\
        & = \alpha \sum_{x, C \in D}\log P_\theta(C \vert x) + \beta \sum_{x, y \in D} \log P_\theta(y \vert x, \hat{C}) \label{eq:loss_components}
    \end{align}}
    where $\log P_\theta(C \vert x)$ optimizes the model to generate English reasoning traces given input in different languages, and $\log P_\theta(y \vert x, \hat{C})$ with $\hat{C} = \theta(x)$ pushes the model to generate final response based on both the input (in target language) and the generated reasoning trace. Additionally, each objective can be balanced with hyperparameter weights $\alpha$ and $\beta$. By default, we set both the hyperparameters $\alpha$ and $\beta$ to $1.0$.

    Effectively, this training approach allows the model to ``reason'' in English (through forcing the ground-truth CoT via the loss function $\log P_\theta(C \vert x)$), a language where it has been extensively trained for reasoning. In other words, this enhances the representation alignment between input prompts in different languages, where they will lead to the same traces of reasoning. The training method also allows the model to understand and provide responses in the low-resource target language. This not only boosts performance on low-resource languages but also simplifies the training process by reducing the need for large amount of data in those languages. In Figure~\ref{fig:loss_comparison}, we visualize the changes of the two loss components in Eq.~(\ref{eq:loss_components}) over the first training epoch. The figure indicates that our English-Pivoted CoT training makes it easier for the model to maintain reasoning in English, as evidenced by the lower initial reasoning loss and slower subsequent decrease. The increasing gap between reasoning and response losses shows our method effectively separates English reasoning from target-language generation, enabling better understanding of the low-resource language. In contrast, Native CoT starts with higher initial reasoning loss that decreases rapidly, resulting in a smaller gap between the two losses, indicating less separation between language understanding and reasoning processes.

    This effect can also be seen in Figure~\ref{fig:weight_changes}, comparing the parameter update patterns of Native CoT training (multilingual reasoning fine-tune with both CoT traces and answers in target language) and our proposed English-Pivoted CoT training for r1-distill-Llama-8B. The figure illustrates key differences in adaptation dynamics. Firstly, considering the absolute magnitude of parameter changes, Native CoT introduces larger updates throughout the model (approximately 1.3 times larger), suggesting a more significant departure from the baseline model’s parameters, which can lead to substantially different reasoning behavior. Secondly, examining the relative changes across model layers, English-Pivoted CoT also concentrates updates in the first and last few layers, particularly within attention-related matrices (highlighted in red boxes). These layers have been shown in prior research to be in charge of language understanding and generation tasks~\cite{tran-etal-2024-irish,wendler-etal-2024-llamas}. Consequently, our method strategically targets these layers, enabling the model to effectively understand and generate final responses in the low-resource language without deviating from the original model’s reasoning capability.

\section{Experiments}
    \subsection{Experiment settings}
        \textbf{Baselines.} We select r1-distill-Llama-8B~\cite{deepseekai2025deepseekr1incentivizingreasoningcapability} and DeepHermes-3-Llama-3-8B~\cite{deephermes3} due to their state-of-the-art performance and open-source availability. We perform fine-tuning on these models following our proposed approach (denote \emph{English-Pivoted CoT Training}). We compare the performance of our method against several existing multilingual adaptation techniques, including few-shot prompting~\cite{shi2023language}, multilingual reasoning fine-tuning (denote \emph{Native CoT Training})~\cite{anand2024multilingualmathematicalreasoningadvancing,lai-nissim-2024-mcot}, model merging~\cite{yadav2023tiesmerging,tao-etal-2024-unlocking} and layer swap~\cite{bandarkar2025layer} with Llama-3.1-8B-Instruct~\cite{grattafiori2024llama3herdmodels} following the default hyperparameter setup.

        Given the limited availability of reasoning data in low-resource languages, following recent works~\cite{tao-etal-2024-unlocking,liu-etal-2025-translation}, we use NLLB-3.3B~\cite{costa2022no} to translate existing reasoning datasets: Bespoke~\cite{bespoke_stratos} into Irish, pensez~\cite{ha2025pensezdatabetterreasoning} into French. For Chinese, a high-resource language, we leverage congliu~\cite{Chinese-Data-Distill-From-R1}, and translate to English for our approach. We sample the amount of data in target language to be $N=8{,}000$ samples, and add in 8{,}000 English samples from the same source to prevent forgetting. In our approach, reasoning traces originally in target languages are translated into English for training purposes. While the amount of data we leverage for adaptation to low-resource languages is relatively small (e.g., compared to 800{,}000 samples used in~\cite{deepseekai2025deepseekr1incentivizingreasoningcapability}), our evaluation results illustrate a clear improvement, highlighting its effectiveness. We note that all evaluation datasets are created or manually verified by humans, ensuring accuracy and quality.

        \textbf{Training setup.} Our training implementation employs HuggingFace Transformers and DeepSpeed. To manage memory constraints efficiently, we set the maximum input sequence length to 16,384 tokens. Models are trained using the AdamW optimizer~\cite{loshchilov2018decoupled} for 3 epochs, with a maximum learning rate of $1\times10^{-5}$. Training is distributed across two Nvidia A100 GPUs (80GB each), with a total batch size of 24.

        \begin{table*}[ht]
        \centering
        \resizebox{0.85\linewidth}{!}{
        \begin{tabular}{@{}lP{5.9cm}ccc@{}}
        \toprule
        \textbf{Model} & \textbf{Setting} & \textbf{AIME2024 (en)} & \textbf{AIME2024 (ga)} & \textbf{LC2024 (ga)} \\ \midrule
        r1-distill-Llama-8B & Zero-shot Prompting & 43.33  & 6.67 & 63.64 \\
        r1-distill-Llama-8B & Few-shot Prompting - Irish CoT & 46.67 & 10.00 &	45.45 \\
        r1-distill-Llama-8B & Few-shot Prompting - English CoT & 46.67	& 6.67	& 32.73 \\ \hline
        r1-distill-Llama-8B & TIES-merging~\cite{yadav2023tiesmerging,tao-etal-2024-unlocking} & 26.67 & 3.33 & 34.55 \\
        r1-distill-Llama-8B & Layer Swap~\cite{bandarkar2025layer} & 36.67 & 16.67 & 30.91 \\
        \hline
        r1-distill-Llama-8B & Native CoT Training~\cite{anand2024multilingualmathematicalreasoningadvancing,lai-nissim-2024-mcot} & 48.33  & 21.67 & 37.14 \\
        \hline
        r1-distill-Llama-8B  &  English-Pivoted CoT Training (OURS) & \textbf{53.33} & \textbf{35.00} & \textbf{73.33} \\
        \midrule
        DeepHermes-3-Llama-3-8B & Zero-shot Prompting & 1.67 & 6.67 & 52.73 \\
        DeepHermes-3-Llama-3-8B & Few-shot Prompting - Irish CoT & 6.67 & 0.00 &	23.64 \\
        DeepHermes-3-Llama-3-8B & Few-shot Prompting - English CoT & 6.67 &	3.33 &	32.73 \\
        \hline
        DeepHermes-3-Llama-3-8B & TIES-merging~\cite{yadav2023tiesmerging,tao-etal-2024-unlocking} & 3.33 & 0.00 & 14.55 \\
        DeepHermes-3-Llama-3-8B & Layer Swap~\cite{bandarkar2025layer} & 3.33 & 3.33 & 25.45 \\ \hline
        DeepHermes-3-Llama-3-8B & Native CoT Training~\cite{anand2024multilingualmathematicalreasoningadvancing,lai-nissim-2024-mcot} & 1.67 & 8.33 & 40.01 \\
        \hline
        DeepHermes-3-Llama-3-8B  &  English-Pivoted CoT Training (OURS) & \textbf{5.00} & \textbf{10.00} & \textbf{54.55} \\ 
        \bottomrule
        \end{tabular}
        }
        \caption{Performance comparison of our English-Pivoted CoT Training approach against baseline methods across English (en) and Irish (ga) reasoning benchmarks. Bold scores indicate the best performance per benchmark.}
        \label{tab:main_result}
        \end{table*}

        \begin{table*}[ht]
        \centering
        \resizebox{0.85\linewidth}{!}{
        \begin{tabular}{@{}llcc@{}}
        \toprule
        \textbf{Model} & \textbf{Setting} & \textbf{concepts \& skills} & \textbf{contexts \& applications} \\ \midrule
        r1-distill-Llama-8B & Zero-shot Prompting & \textbf{84.85} & 31.82 \\
        r1-distill-Llama-8B & Native CoT Training & 47.50 & 31.82 \\
        \hline
        r1-distill-Llama-8B  &  English-Pivoted CoT Training (OURS) & 82.82 & \textbf{59.09} \\
        \midrule
        DeepHermes-3-Llama-3-8B & Zero-shot Prompting & 57.58 & 45.45 \\ 
        DeepHermes-3-Llama-3-8B & Native CoT Training & 57.58 & 13.64 \\ 
        \hline
        DeepHermes-3-Llama-3-8B &  English-Pivoted CoT Training (OURS) & \textbf{57.58} & \textbf{50.00} \\ \bottomrule
        \end{tabular}
        }
        \caption{Comparisons of low-resource language understanding on LC2024, where the exam is split into 2 parts: concepts \& skills and contexts \& applications. The latter part has additional contextual and real-world descriptions, requiring a higher level of Irish language understanding to comprehend the input. Bold scores indicate the best performance per benchmark.}
        \label{tab:lc2024_result}
        \end{table*}

        \textbf{Benchmarks.} We evaluate the reasoning capabilities of all models on standard English benchmarks, including AIME2024~\cite{aime_1983_2024} (challenging Olympiad-level math) and MGSM~\cite{shi2023language} (high school-level math). Additionally, we use MGSM for evaluating performance in French and Chinese (as it is a multilingual benchmark dataset), supplemented by MATH-hard~\cite{openllm-French-leaderboard} 
        for French, follow the French LLM leaderboard~\cite{openllm-French-leaderboard}. For the Irish language, due to lack of existing datasets, we contribute two new ones:

        \begin{itemize}
            \item Irish version of AIME2024: translated and verified by two Irish speakers.
            \item Leaving Certificate 2024 Math exam - Higher Level (LC2024): derived from official Irish Leaving Certificate~\cite{wikipedia_leaving_certificate_ireland} examination with 2 splits: concepts\&skills and contexts\&applications/ We extract individual questions that have concrete answers (excluding proof-based questions) and do not require modalities beyond text (e.g., geometric diagrams). The questions are formatted in LaTeX with a total of 55 samples.
        \end{itemize}

         We adopt each benchmark’s default hyperparameters, leveraging a decoding temperature of $0.6$, top-k sampling with $k = 0.95$, limiting generation outputs to 16{,}384 tokens.

        \begin{figure*}[ht]
        \centering
        \includegraphics[width=0.9\linewidth]{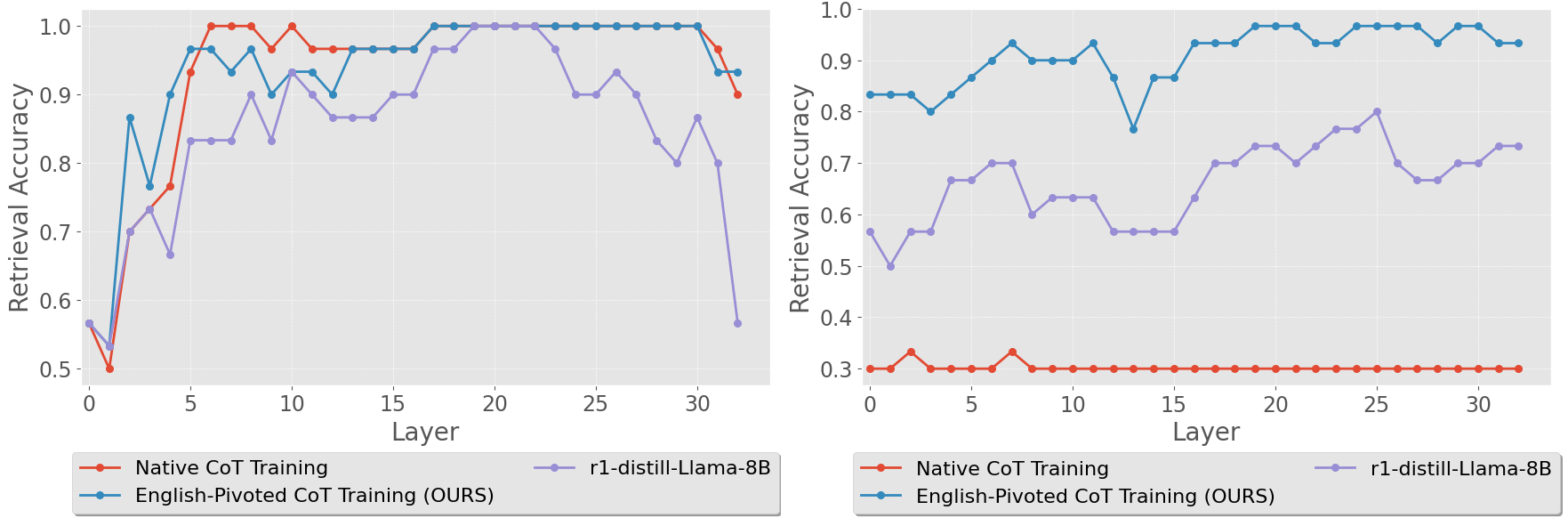}
        \caption{Representation retrieval accuracy between \emph{Left}: questions, and \emph{Right}: questions and generated CoT traces of the same questions in different languages.}
        \label{fig:aime_retrieval}
        \end{figure*}

    \subsection{Results and analysis}
        \label{sec:results}

        \textbf{Reasoning capability can be effectively transferred to low-resource languages.} Table~\ref{tab:main_result} presents the performance of models trained with our approach (English-Pivoted CoT Training) compared to baselines on English and Irish benchmarks. 
        Our method demonstrates a strong improvement: on the Irish version of AIME2024, while all baselines perform poorly, our approach obtains a clear gap of upto 28.33\% on r1-distill-Llama-8B baseline. More specifically, we achieve accuracies of 35.00\% and 10.00\% when fine-tuned on r1-distill-Llama-8B and DeepHermes-3-Llama-3-8B, respectively, matching performance on the English version. On LC2024, which is relatively less challenging than AIME2024, although r1-distill-Llama-8B achieves an acceptable accuracy of 63.64\%, our approach surpasses it by a 10\% margin (73.33\%). Further inspection shows that both baseline models generate all CoTs and final answers in English, regardless of prompting in which languages. This default behavior aligns with the English prompting approaches in~\cite{qin-etal-2023-cross,shi2023language}. Interestingly, performances on LC2024 (authored by Irish speakers) drops compared to the translated Irish AIME2024 for all baseline approaches. We hypothesize the model relies on familiar math notation (same across all languages) to infer the task and then applies English reasoning, whereas interventions such as prompting, CoT fine‑tuning or model merging disrupt this implicit strategy, causing parameter interference in reasoning in an extremely low-resource language.


        \textbf{Improved language understanding capability.} To investigate whether our approach also enhances language understanding or simply overfits to mathematical phrases, we split LC2024 into its two original sections:  \emph{concepts \& skills} and \emph{contexts \& applications} - with more contextual and real-world application descriptions, requires deeper Irish language understanding. Due to space limitation, we present the results of our method and 2 strongest baselines, with more results in the Appendix. As shown in Table~\ref{tab:lc2024_result}, Native CoT training reduces the model’s performance for DeepHermes-3-Llama-3-8B from 45.45 to 13.64 on \emph{concepts \& skills}, while maintaining the same performance on r1-distill-Llama-8B. This shows that training to force the model to reason in low-resource languages affects its ability to generate effective CoTs. In contrast, our approach yields substantially gain in \emph{contexts \& applications}, indicating that not only does it enable effective reasoning in low-resource languages but also improves language understanding.

        \begin{table}[!t]
        \centering
        \resizebox{1.0\linewidth}{!}{
        \begin{tabular}{@{}P{2.6cm}P{1.8cm}C{1.3cm}C{2.25cm}C{2.5cm}@{}}
        \toprule
        \textbf{Training Setting} & \textbf{Prompting Setting} & \textbf{AIME2024} & \textbf{LC2024 - concepts\&skills} & \textbf{LC2024 - contexts\&applications} \\ \midrule
        Mixed CoT Lang. & Irish CoT & 33.33 &	3.33 &	54.55  \\ 
        Mixed CoT Lang. & English CoT & 33.33 & 30.00 & 61.82 \\ 
        \midrule
        Two-stage  & Irish CoT  & 40.00 &	6.67 &	45.45 \\ 
        Two-stage & English CoT & 33.33 &	33.33 &	58.18 \\
        \bottomrule
        \end{tabular}
        }
        \caption{Irish task performance by CoT language for r1-distill-Llama-8B trained on English and Irish CoTs.}
        \label{tab:language_of_reasoning_result}
        \end{table}

        \textbf{Diverged reasoning traces when forced to reason in different languages.} 
        We assess cross‑language alignment on AIME2024 by retrieving English embeddings for Irish questions (with and without CoTs) using average token representations across layers. As illustrated in Figure~\ref{fig:aime_retrieval}, target‑language‑only models show near‑perfect question‑only alignment (up to 100\%), but this falls to around 35\% once CoTs are included, revealing divergent internal representations. Figure~\ref{fig:example} illustrates how minor embedding shifts (e.g., confusing “independence” and “dependence”) can alter interpretation. In contrast, English‑Pivoted CoT Training maintains almost 100\% alignment for both questions and reasoning traces, demonstrating stable, language‑agnostic embeddings.

        Additionally, we experience with \textbf{controlling the CoT language at test time}. We introduce 2 additional training settings: 1) Mixed CoT Language: 50\% of CoTs in Irish, 50\% in English, 2) Two‑Stage: first stage uses English‑Pivoted CoT to align comprehension, then a second stage fine‑tunes on native Irish CoTs to refine reasoning in the target language. We prepend each rationale with ``Let's think in X language'' and fix this prompt at inference to control the CoT language. Table~\ref{tab:language_of_reasoning_result} reports that on Irish AIME2024 (olympiad‑level), all methods perform similarly, but English‑CoT prompted models clearly outperform on both LC2024 subsets. This suggests that in data‑limited scenarios, directing model to think in English remains the most effective strategy for robust multilingual reasoning.

        \begin{figure}[t]
        \centering
        \includegraphics[width=0.95\linewidth]{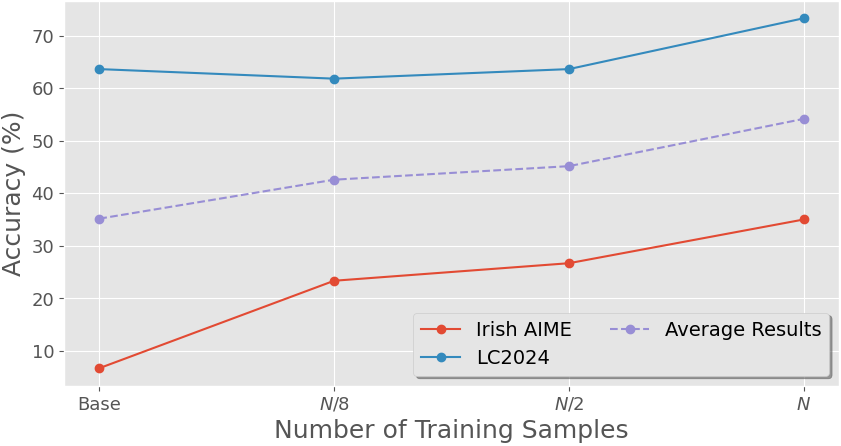}
        \caption{Performance of English-Pivoted CoT training on Irish AIME and LC2024 with varying number of training samples on r1-distill-Llama-8B (default is $N$).}
        \label{fig:varying_samples}
        \end{figure}

        \begin{figure}[!t]
        \centering
        \includegraphics[width=0.95\linewidth]{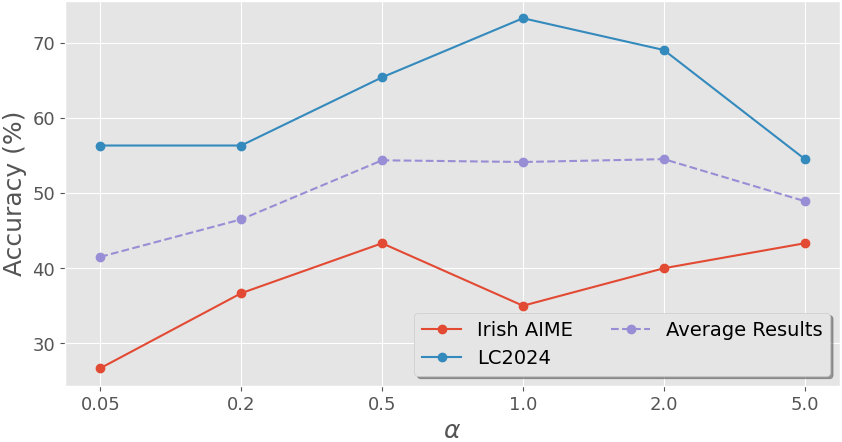}
        \caption{Performance of English-Pivoted CoT training with different $\alpha$ (default: $1.0$) during training ($\beta=1.0$) on r1-distill-Llama-8B. x-axis is evenly spaced for visualization.}
        \label{fig:varying_alpha}
        \end{figure}

        \begin{table*}[!t]
        \centering
        \resizebox{0.85\linewidth}{!}{
        \begin{tabular}{@{}llC{1.6cm}C{1.4cm}C{1.6cm}C{1.4cm}@{}}
        \toprule
        \textbf{Model} & \textbf{Setting} & \textbf{AIME2024 (en)}& \textbf{MSGM (en)} & \textbf{MATH-hard (fr)} & \textbf{MSGM (fr)} \\ \midrule
        r1-distill-Llama-8B & Zero-shot Prompting & 43.33 & 79.6 & 49.74 & 54.8 \\
        r1-distill-Llama-8B & Native CoT Training & 45.00 & 82.0 & 31.90 &  61.2 \\
        \hline
        r1-distill-Llama-8B  &  English-Pivoted CoT Training (OURS) & \textbf{50.00} & \textbf{89.6} & \textbf{70.01} & \textbf{83.2} \\ 
        \midrule
        DeepHermes-3-Llama-3-8B & Zero-shot Prompting & 1.67 & \textbf{85.2} & 3.69 & 21.6 \\
        DeepHermes-3-Llama-3-8B & Native CoT Training & 3.33 & 70.8 & 6.26 & 44.8 \\
        \hline
        DeepHermes-3-Llama-3-8B  &  English-Pivoted CoT Training (OURS) & \textbf{8.33} & 76.8 & \textbf{32.27} & \textbf{77.2} \\
        
        \bottomrule
        \end{tabular}
        }
        \caption{Ablation study of fine-tuning on French - a medium resource language, and benchmark across English and French reasoning datasets. Bold scores indicate the best performance per benchmark.}
        \label{tab:main_result_fr}
        \end{table*}

        \begin{table*}[!t]
        \centering
        \resizebox{0.85\linewidth}{!}{
        \begin{tabular}{@{}llccc@{}}
        \toprule
        \textbf{Model} & \textbf{Setting} & \textbf{AIME2024 (en)} & \textbf{MSGM (en)} & \textbf{MSGM (zh)} \\ \midrule
        r1-distill-Llama-8B & Zero-shot Prompting & 43.33 & 79.6 & 65.2 \\
        r1-distill-Llama-8B & Native CoT Training & 46.33 & 92.4 & \textbf{81.6} \\
        \hline
        r1-distill-Llama-8B  &  English-Pivoted CoT Training (OURS) & \textbf{56.67} & \textbf{92.4} & 70.0 \\ 
        \midrule
        DeepHermes-3-Llama-3-8B & Zero-shot Prompting & 1.67 & 85.2 & 50.8 \\
        DeepHermes-3-Llama-3-8B & Native CoT Training & 11.67 & 83.2 & 61.6 \\
        \hline
        DeepHermes-3-Llama-3-8B &  English-Pivoted CoT Training (OURS) & \textbf{13.33} & \textbf{89.6} & \textbf{62.4} \\ 
        \bottomrule
        \end{tabular}
        }
        \caption{Ablation study of fine-tuning on Chinese - a high resource language, where baselines have high level of understanding and tend to output CoT traces in Chinese. Bold scores indicate the best performance per benchmark.}
        \label{tab:main_result_zh}
        \end{table*}

        \textbf{Sample efficiency of English-Pivoted CoT Training.} 
        By decoupling reasoning from language learning, our method needs only comprehension fine‑tuning. As Figure~\ref{fig:varying_samples} shows, with just 1{,}000 samples ($N/8$) it beats the base model by over 5\%, and accuracy climbs further with more data. This highlights its strong performance in low‑resource settings.

        \textbf{Learning dynamic of English-Pivoted CoT Training.} 
        Varying the CoT‑loss weight $\alpha$ (with $\beta = 1$) in Eq.~\ref{eq:loss_components} (Figure~\ref{fig:varying_alpha}) reveals that both reducing and over‑emphasizing $\alpha$ degrade performance. This highlights the need to balance the English‑CoT and target‑response objectives.
        These results underscore the need to balance the reasoning trace and final response objectives in English‑Pivoted CoT Training.

        \begin{figure}[t]
        \centering
        \includegraphics[width=0.95\linewidth]{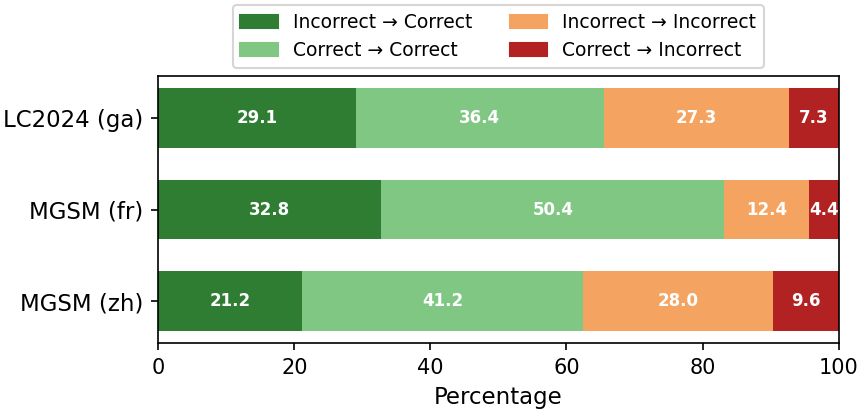}
        \caption{Changes in correction rate from r1-distill-Llama-8B when finetuned with our proposed approach across 3 data resources, low (Irish), medium (French), and high (Chinese).}
        \label{fig:correction_rate}
        \end{figure}

            \textbf{Generalizability to medium- and high-resource languages.} 
        Experiments on Chinese and French - medium to high-resource languages~\cite{11234/1-5787,chang-etal-2024-multilinguality}, present in Table~\ref{tab:main_result_fr} and Table~\ref{tab:main_result_zh}, indicate the generalizability: 
        on French MGSM, r1‑distill‑Llama‑8B scores 54.8\%, while our approach boosts it to 83.2\%; on Chinese, English pivoting yields only marginal gains over baseline 65.2\%, likely because the model naturally reasons in Chinese, so forcing English reasoning can cause interference.
        Furthermore, the percentage of the new model corrects an incorrect answer by the base model and vice versa are shown in Figure~\ref{fig:correction_rate}.
        We can see that on MGSM (zh), the correct $\rightarrow$ incorrect rate is the highest, reflecting these conflicts.
        Nevertheless, our method still achieves an average improvement of 14.30\% across the 3 resource regimes (low, medium, and high-resources), demonstrating its broad applicability.

\section{Conclusion}
    In this work, we propose a novel approach named English-Pivoted CoT Training to effectively transfer the reasoning capabilities of LLMs to low-resource languages. Through training to explicitly aligning reasoning processes across languages by forcing CoT traces to be in the dominant language, our approach achieves significant performance gains (up to 28.33\%) over existing techniques, and is generalizable to other resource regimes (medium and high-resource languages). Furthermore, our analysis provides insights into multilingual LLM behavior, particularly the benefit of separating language understanding from reasoning. By contributing the first-ever Irish mathematical reasoning benchmark (LC2024), we also aim to support future research in multilingual reasoning. 
    Future work can be interested in extending to other low-resource languages and tasks, and developing strategies to further enhance cross-lingual reasoning.

\section*{Ethics Statement}
    Our work advances language technologies for low-resource and endangered languages, with a particular focus on Irish. By leveraging English-Pivoted CoT Training, we aim to enhance the accessibility and usability of these languages, thereby promoting linguistic diversity and preserving cultural heritage.

    All training and evaluation data are drawn from publicly available sources. Our newly released LC2024 dataset is based on Regulations on Re-use of Public Sector Information 2005, SI 279 of 2005, and is licensed for non-commercial research and educational use only.

\section*{Acknowledgments}

We would like to acknowledge CloudCIX Limited for the generous collaborative support of computing resources on their NVIDIA HGX/H100 GPU cluster. This publication has emanated from research supported in part by grants from Research Ireland under Grant [12-RC-2289-P2] and [18/CRT/6223] which is co-funded under the European Regional Development Fund. For the purpose of Open Access, the author has applied a CC BY public copyright licence to any Author Accepted Manuscript version arising from this submission.

\bibliography{aaai2026}

\end{document}